\documentclass[10pt,twocolumn,letterpaper]{article}

\usepackage[pagenumbers]{cvpr} 

\usepackage[utf8]{inputenc}
\usepackage{kotex}
\usepackage{xcolor}

\definecolor{cvprblue}{rgb}{0.21,0.49,0.74}
\usepackage[pagebackref,breaklinks,colorlinks,allcolors=cvprblue]{hyperref}

\def\paperID{25} 
\def\confName{CVPR}
\def\confYear{2025}

\title{Comparison Reveals Commonality: Customized Image Generation through Contrastive Inversion}

\author{
    Minseo Kim\textsuperscript{1}, 
    Minchan Kwon\textsuperscript{1}, 
    Dongyeun Lee\textsuperscript{1}, 
    Yunho Jeon\textsuperscript{2}\textsuperscript{\textdagger}, 
    and Junmo Kim\textsuperscript{1}\textsuperscript{\textdagger} \\
    \textsuperscript{1}Korea Institute of Science and Technology, South Korea \\
    \textsuperscript{2}Hanbat National University, South Korea \\
    {\tt\small \{alstj1571,kmc0207,ledoye\}@kaist.ac.kr, yhjeon@hanbat.ac.kr, junmo.kim@kaist.ac.kr}
}

\begin{document}
\maketitle
\footnotetext{\textsuperscript{\textdagger}Joint corresponding authors.}

\begin{abstract}

The recent demand for customized image generation raises a need for techniques that effectively extract the common concept from small sets of images. 
Existing methods typically rely on additional guidance, such as text prompts or spatial masks, to capture the common target concept. 
Unfortunately, relying on manually provided guidance can lead to incomplete separation of auxiliary features, which degrades generation quality.
In this paper, we propose \textbf{Contrastive Inversion}, a novel approach that identifies the common concept by comparing the input images without relying on additional information. 
We train the target token along with the image-wise auxiliary text tokens via contrastive learning, which extracts the well-disentangled true semantics of the target. 
Then we apply disentangled cross-attention fine-tuning to improve concept fidelity without overfitting. 
Experimental results and analysis demonstrate that our method achieves a balanced, high-level performance in both concept representation and editing, outperforming existing techniques.
\end{abstract}

\section{Introduction}
\label{sec:intro}

Recent advances in text-to-image (T2I) diffusion models \cite{ldm,dalle2,imagen} have shown impressive capabilities to generate images that faithfully reflect text descriptions.
Building on this success, there is an increasing demand for customized image generation \cite{textual_inversion,dreambooth,custom_diffusion,han2023svdiff,hyperdreambooth,jia2023taming,wei2023elite,bas,dco,detailedprompt1,nam2024dreammatcher,cones2}, where unique concepts like “my dog” or “my drawing style” are learned from a few user-provided images.

Most existing methods are based on Textual Inversion \cite{textual_inversion}, which learns the latent embedding of a special text token to represent a shared concept across a limited set of images. 
However, due to the small amount of training data, the model often overfits by memorizing not only the target concept but also irrelevant “auxiliary” concepts (e.g., background or pose). 
Consequently, the model ends up merely reproducing the input images, rather than editing them according to the user's instructions.
This issue stems from the confusion between target and auxiliary concepts, highlighting the importance of concept disentanglement.

\begin{figure}
  \centering
  \includegraphics[width=1.0\linewidth]{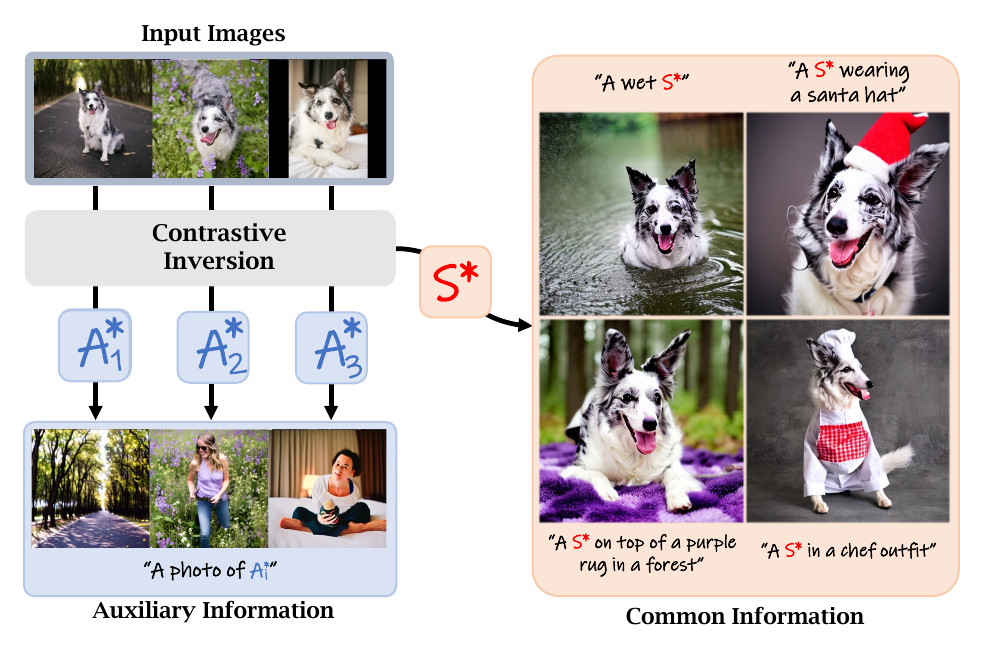}
  \caption{\textbf{Overview of Contrastive Inversion.} From the input images, we separate the commonalities (\(S^*\)) and the image-specific auxiliary (\(A_i^*\)) information into text tokens without additional guidance. Only \(S^*\) is used in the generation with text prompts.}
  \label{fig:title}
\end{figure}

\begin{figure*}[t]
  \centering
  \includegraphics[width=1.0\linewidth]{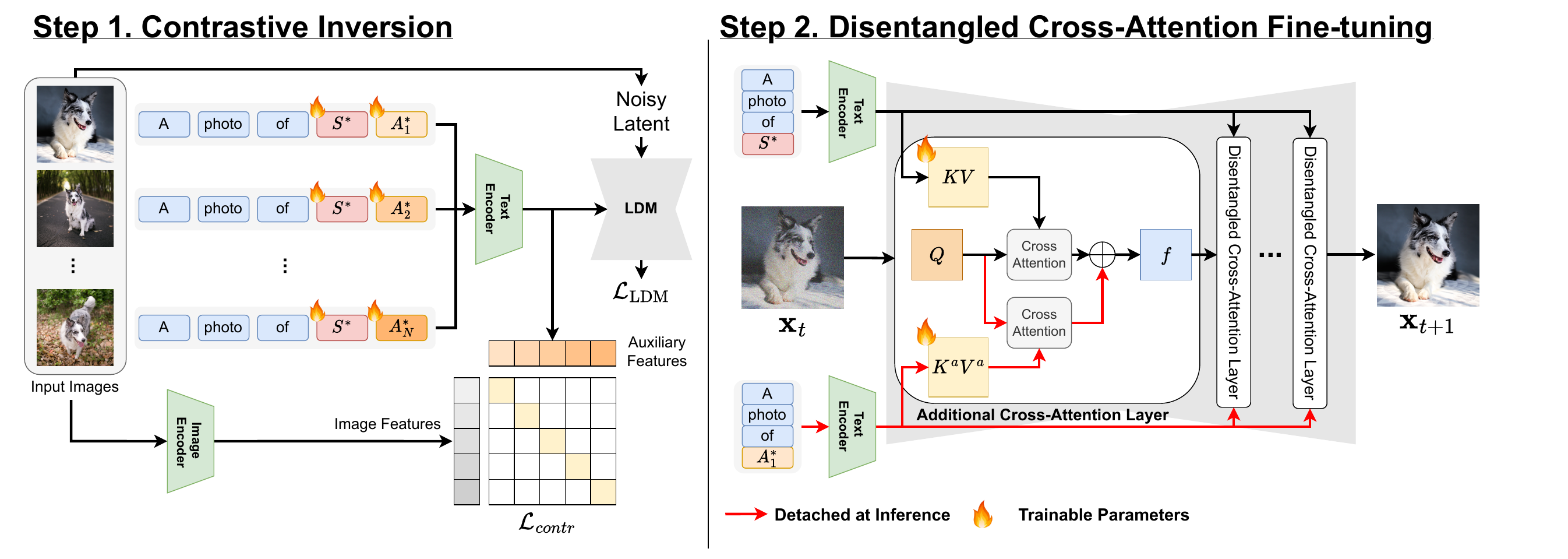}
  \caption{\textbf{Two-stage training pipeline of our method.} In Step~1, we optimize both the target and the image-wise auxiliary tokens to represent their respective visual concepts, effectively disentangling the target concept via contrastive learning. In Step~2, we reuse the learned tokens to fine-tune the diffusion model. Auxiliary tokens carrying irrelevant information are treated in additional cross-attention layers, which are discarded in inference.}
  \label{fig:method}
\end{figure*}

Previous approaches have sought to achieve concept disentanglement by incorporating additional guidance.
Early methods like DreamBooth \cite{dreambooth} and Custom Diffusion \cite{custom_diffusion} indicate the target concept by assigning a class noun (e.g., ``dog") to the target token.
Subsequent works introduced more explicit guides, such as object masks~\cite{jia2023taming,wei2023elite,bas,han2023svdiff}, detailed text captions~\cite{dco,detailedprompt1}, or attention control~\cite{han2023svdiff,nam2024dreammatcher,cones2}.
For instance, ELITE~\cite{wei2023elite} treats intermediate features from the CLIP~\cite{clip} image encoder as auxiliary information and discards them during inference.
Similarly, DisenBooth~\cite{disenbooth} utilizes the CLIP image embedding but masks the portion corresponding to the class noun to construct its auxiliary information. 
However, the effectiveness of these methods is fundamentally limited by the quality of the manual guidance, as neither CLIP embeddings nor class nouns can fully capture fine-grained visual details.

To overcome this limitation, we propose \textbf{Contrastive Inversion}, a novel fine-tuning scheme for customized image generation that achieves concept disentanglement by comparing the given images.
We jointly optimize image-wise auxiliary text tokens alongside the target text token via contrastive learning, eliminating the need for additional guidance.
Our setup distributes the irrelevant visual semantics of each image into the auxiliary tokens, encouraging the target token to capture only the intended concept.

Following the inversion, we introduce \textbf{Disentangled Cross-attention Fine-tuning} to faithfully integrate the learned tokens into the diffusion model while preventing overfitting.
The auxiliary tokens are inserted into separate cross-attentional layers, allowing those layers to be removed from the model after training.
This straightforward yet effective technique prevents overfitting to the auxiliary information, improving both concept fidelity and editability.
Experimental results demonstrate that our method outperforms existing techniques across a wide range of concepts.

\section{Methods}

\subsection{Contrastive Inversion of Auxiliary Tokens}

\paragraph{Image-wise auxiliary tokens}
Given \(N\) training images $\mathbf{x}_i$ for $i=1\dots N$, our goal is to encode shared information of the images into a unique text token \(S^*\). 
To help concept disentanglement, we introduce image-wise auxiliary tokens \(A^*_i\) for each image \(\mathbf{x}_i\) and jointly optimize with \(S^*\).

The tokens are updated by following LDM~\cite{ldm} loss : 
\begin{equation}
     \mathcal{L}_{\mathrm{LDM}} := \mathbb{E}_{\mathbf{z}, \epsilon,y, t}[w_{t}\|\epsilon-\epsilon_{\theta}(\mathbf{z}_{t},\tau_T(y),t)\|^2_2],
 \label{eq:ldm_loss}
\end{equation}
\vspace{10pt}
where \(\epsilon\sim \mathcal{N}(0,I)\) is random noise, \(\mathbf{z}_t\sim \mathcal{E}(\mathbf{x})\) is the latent image at time step \(t\), and \(\tau_T\) is a CLIP~\cite{clip} text encoder. 
The function \(\epsilon_{\theta}\) is the noise prediction network.
We construct the text condition y as “A photo of \(S^*\,A_i^*\),” providing both target and auxiliary tokens.
We freeze the model weights and optimize only the tokens.

\begin{figure*}[t]
  \centering
  \includegraphics[width=1\linewidth]{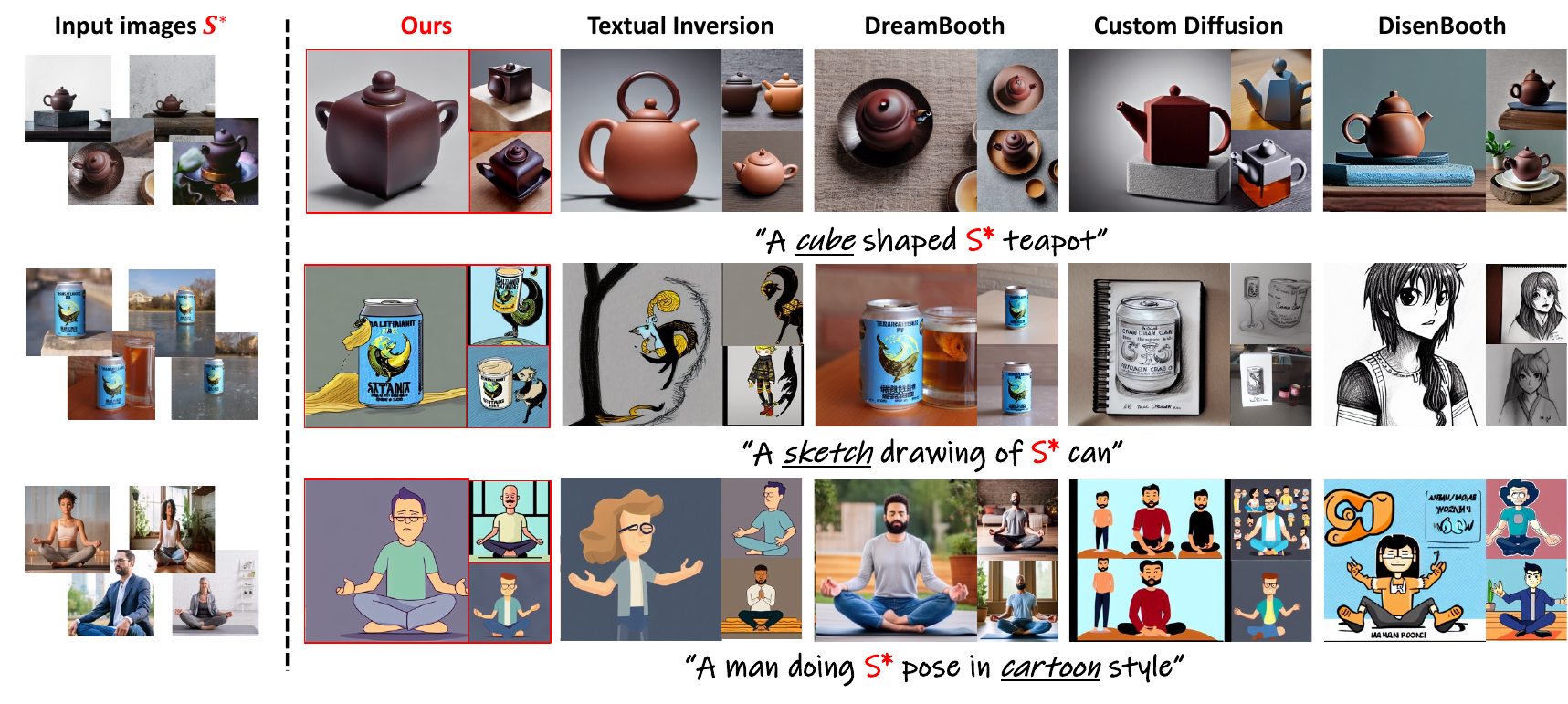}
  \caption{\textbf{Qualitative results.} Based on a precise understanding of the target concept, our method maintains high concept fidelity while integrating with given text prompts. In contrast, previous methods exhibit severe overfitting or underfitting, and struggle to fuse the learned concept with different text conditions.}
  \label{fig:qualitative}
\end{figure*}

\paragraph{Contrastive Inversion}
Optimizing tokens with only the LDM loss can lead to a trivial solution where the auxiliary tokens absorb all features, leaving the target token \(S^*\) meaningless.
To ensure the target concept is encoded exclusively in \(S^*\), we introduce InfoNCE loss~\cite{oord2018representation} on auxiliary tokens.

Each image and its corresponding auxiliary token are treated as a positive pair, while all other image-token pairs are considered negative.
We optimize tokens to attract positive pairs and repel negative ones in the CLIP embedding space, where the distance is measured by cosine similarity.

Formally, the contrastive objective is defined as:

\begin{equation}
\mathcal{L}_{\mathrm{contr}} := \sum_{i=1}^{N} -\log \left( \frac{\exp(\mathrm{sim}(\tau_T(A^*_i),\tau_I(\mathbf{x}_i)))}{\sum_{j=1}^{N}\exp(\mathrm{sim}(\tau_T(A^*_i),\tau_I(\mathbf{x}_j))) } \right),
\label{eq:contr_loss}
\end{equation}
\vspace{8pt}

where $\tau_T$and $\tau_I$ are fixed CLIP text and image encoders.

Contrastive objective encourages \(A_i^*\) to learn only the distinctive features of its particular image by comparing each auxiliary token to all other images, achieving the high-level feature disentanglement.
By jointly optimizing the contrastive loss with LDM loss, we obtain optimal representation vectors where the target token captures the common concept and the auxiliary tokens learn image-specific features.


\subsection{Disentangled Cross-attention Fine-tuning}

While Contrastive Inversion finds text tokens that effectively disentangle the target and auxiliary concepts, generating images with high fidelity remains challenging with a frozen generative model.
To enhance concept fidelity, we further fine-tune the key (K) and value (V) projection matrices within the cross-attention layers of the diffusion model, which are responsible for injecting the textual condition.

However, we observed that the fine-tuning process can lead to overfitting, causing the model to memorize both target and auxiliary concepts regardless of the token disentanglement. To mitigate the influence of auxiliary concepts, we duplicate the key and value matrices to be optimized. The original matrices \(K\) and \(V\) are trained exclusively with the target token \(S^*\), while the copied, auxiliary matrices \(K^a\) and  \(V^a\) are trained solely with the optimized auxiliary tokens \(A_i^*\). During training, we sum the outputs of these two independent attention pathways as follows:

\begin{equation}
\label{eq:attention}
\mathbf{f} = \mathrm{Attention}(Q,K,V)\ + \ \lambda\ \mathrm{Attention}(Q,K^{a},V^{a}).
\end{equation}

Upon completion of training, we simply discard the auxiliary matrices \(K^a\) and \(V^a\). This straightforward strategy prevents the auxiliary concepts from being memorized in the final generative model, ensuring it is influenced only by the disentangled target concept. The overall pipeline is illustrated in Figure~\ref{fig:method}.

\section{Experiments}

\subsection{Quantitative Evaluations}

We quantitatively analyze the effectiveness of our method from two perspectives:
\begin{itemize}
\item \textbf{Concept Fidelity}: \textit{Does it accurately preserve the visual details of the target concept?}
\item \textbf{Prompt Alignment}: \textit{Does it faithfully follow the given text prompts?}
\end{itemize}

Most existing methods exhibit a trade-off between these criteria, as prolonged training for higher fidelity causes the model to lose its prior knowledge.
A high score in one metric without a balance in the other suggests severe overfitting or underfitting.
Following the protocol of prior works~\cite{dreambooth, custom_diffusion}, we measure concept fidelity with DINOv2~\cite{dinov2} and prompt alignment with CLIP-T~\cite{clip}, using the DreamBench~\cite{dreambooth} dataset.
Further details on the experimental settings are provided in the Appendix~\ref{appendix:Training Details}.

Table~\ref{tab:main} summarizes our experimental results.
Textual Inversion~\cite{textual_inversion}, which only optimizes a text token without LDM fine-tuning, performs poorly on both metrics.
The widely used DreamBooth~\cite{dreambooth} and Custom Diffusion~\cite{custom_diffusion} exhibit a severe trade-off, indicating overfitting and underfitting, respectively.
In contrast, our method and DisenBooth, both designed for concept disentanglement, achieve a strong balance by securing near-optimal scores on both metrics.
This result demonstrates that effectively disentangling the target concept from auxiliary features is key to improving the trade-off between concept fidelity and prompt alignment.

\subsection{Qualitative Analysis}
\label{sec:qualitative}

Figure~\ref{fig:qualitative} presents a qualitative comparison of diverse object and pose\footnote{\url{https://github.com/bighuang624/ActionBench?tab=readme-ov-file}} datasets. 
DreamBooth accurately represents the given object but completely ignores text prompts such as ``cube", ``sketch", or ``cartoon", instead reproducing the input images. 
Custom Diffusion, on the other hand, tends to over-prioritize the prompt, leading to a significant drop in the target concept's fidelity. 
Among them, our method successfully represents the desired concept with high fidelity while adhering to the given prompt.
Notably, our method is the only one that successfully transforms the given teapot into a cube shape while preserving its original color and design. 
This success is attributable to our model's precise understanding of the target concept within the input images.

Although DisenBooth showed competitive quantitative metrics, Figure~\ref{fig:qualitative} shows its limitations. 
DisenBooth distinguishes between target and auxiliary concepts using a predefined rule based on CLIP image embeddings and a single class noun.
This rule-based approach does not always guarantee a complete understanding of the target concept, failing to fuse the target token with the given prompt. 
In contrast, our method learns token embeddings via contrastive learning, enabling it to accurately capture the true semantics of the target concept.
The difference is particularly evident with prompts that significantly alter the target object's appearance, such as ``cube" or ``sketch".

\begin{table}
\small
\centering
\setlength{\tabcolsep}{6pt}
\begin{tabular}{lcc}
    \toprule
    \textbf{Method} & \textbf{DINOv2$\uparrow$} & \textbf{CLIP-T$\uparrow$} \\ 
    \midrule
    Pre-trained (baseline) & 0.314$ \pm $0.106 & 0.329$ \pm $0.027 \\ 
    \midrule
    Textual Inversion \cite{textual_inversion} & 0.458$\pm$0.107 & 0.297$\pm$0.031 \\ 
    DreamBooth \cite{dreambooth} & \underline{0.599$\pm$0.067} & \textcolor{gray}{0.277$\pm$0.029} \\ 
    Custom Diffusion \cite{custom_diffusion} & \textcolor{gray}{0.457$\pm$0.108} & \underline{0.316$\pm$0.029} \\ 
    DisenBooth \cite{disenbooth} & \textbf{0.530$\pm$0.093} & \textbf{0.301$\pm$0.028} \\ 
    \midrule
    Ours w/o LDM Fine-Tuning & 0.465$\pm$0.115 & 0.304$\pm$0.032 \\ 
    Ours & \textbf{0.530$\pm$0.101} &\textbf{ 0.302$\pm$0.032} \\ 
    \bottomrule
\end{tabular}
\caption{\textbf{Quantitative evaluation on the DreamBench dataset.} We measure concept fidelity (DINOv2) and prompt alignment (CLIP-T) at the best-performing epoch for each method, modified from a pre-trained LDM. The best score for each metric is \underline{underlined}, while the worst is marked in \textcolor{gray}{gray}. Our method demonstrates the best performance when considering both metrics.}
\label{tab:main}
\end{table}

\vspace{-0.1cm}
\subsection{Ablation Studies}
\vspace{-0.1cm}

\paragraph{Auxiliary Token Capacity}

\begin{figure}[t]
  \centering
  \includegraphics[width=1.0\linewidth]{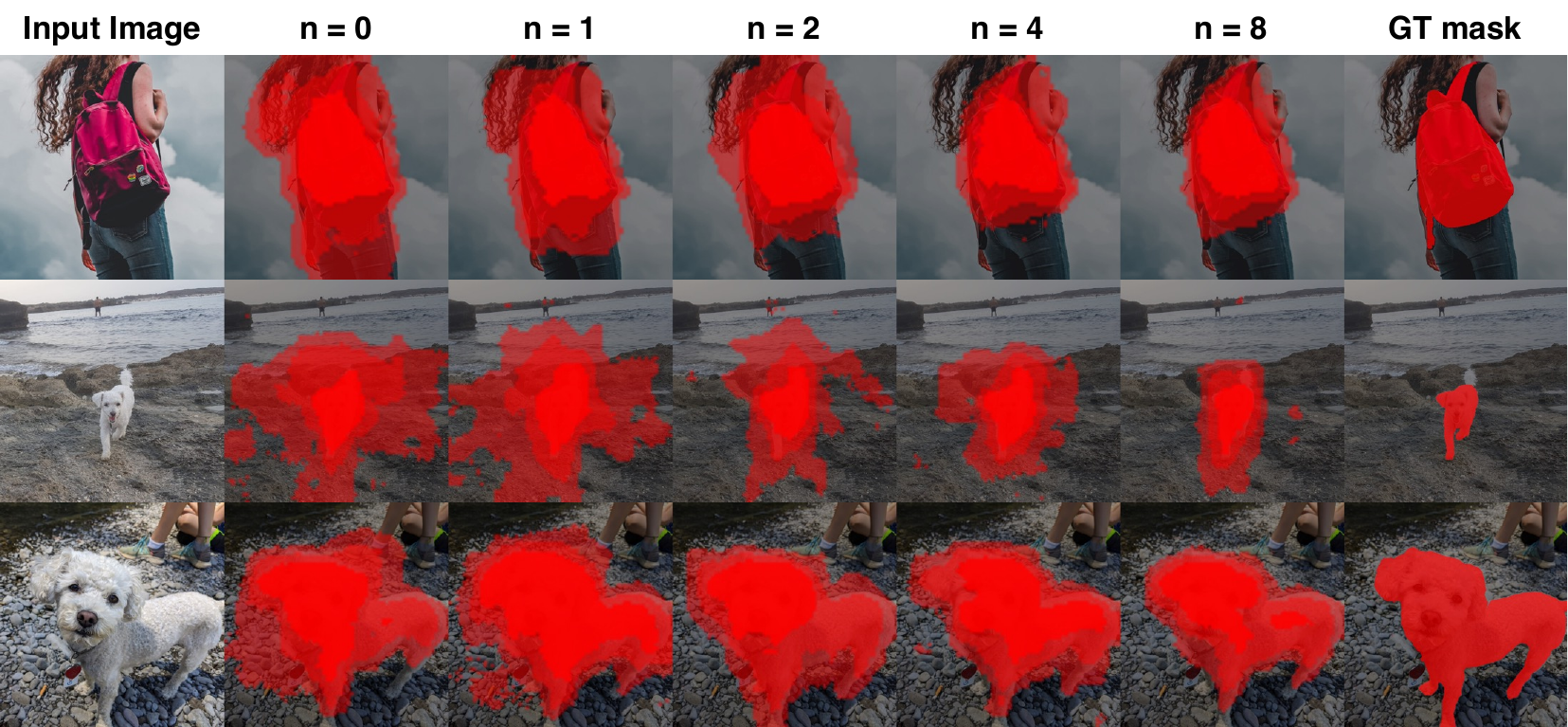}
  \caption{\textbf{Cross-attention maps related to the target token.} $n$ denotes the number of token embeddings assigned to each auxiliary token. As $n$ increases, the target token more precisely represents the target object.}
  \label{fig:attnmap}
\end{figure}

We analyze the effect of auxiliary token capacity on concept disentanglement by visualizing the cross-attention maps of the target token. In Figure~\ref{fig:attnmap}, $n$ denotes the number of token embeddings assigned to a single auxiliary token $A_i^*$. A larger $n$ provides the auxiliary token with greater capacity to absorb image-specific details, thereby helping to isolate the common concept.

\vspace{-2cm}

When $n=0$ (i.e., no auxiliary token), the target token is trained alone and becomes entangled with surrounding objects, such as the woman holding the bag. As $n$ increases, the auxiliary token captures more of these peripheral details, allowing the target token to focus more precisely on the intended object. This result validates our core hypothesis: by providing sufficient capacity for auxiliary tokens to learn irrelevant information, the target token can more accurately represent the desired concept.

\vspace{-0.5cm}
\paragraph{Disentangled Cross-Attention Fine-tuning}

We conduct an ablation study to verify the effectiveness of each stage in our proposed pipeline, with results shown in Table~\ref{tab:main}. The row ``Ours w/o LDM Fine-Tuning" reports the performance after only the first stage, Contrastive Inversion. Compared to standard Textual Inversion, this stage alone improves both concept fidelity (DINOv2) and prompt alignment (CLIP-T). However, since the generative model remains frozen, the concept fidelity score is modest. The final row, "Ours," shows the results after applying our second stage, Disentangled Cross-Attention Fine-tuning. This stage substantially improves the DINOv2 score from 0.465 to 0.530 while the CLIP-T score remains stable. This confirms that our fine-tuning strategy effectively enhances visual fidelity without overfitting to auxiliary concepts, thus preserving editability.
\vspace{-0.6cm}
\section{Conclusion}
\vspace{-0.2cm}
We introduced Contrastive Inversion, a framework that disentangles a target concept from auxiliary features using contrastive learning on text tokens. Combined with our disentangled cross-attention fine-tuning, the method enhances concept fidelity while preventing overfitting.

For future work, we suggest extending our approach from concrete objects to abstract concepts like artistic styles. Another key direction is improving efficiency in terms of training time and memory. The modularity of our method allows for straightforward integration with parameter-efficient fine-tuning (PEFT) techniques to reduce these computational costs. We believe our approach provides a foundational principle for advancing the field of personalized image generation.
{
    \small
    \bibliographystyle{ieeenat_fullname}
    \bibliography{main}
}

\clearpage
\appendix

\section{Training Details} 
\label{appendix:Training Details}
All experiments are conducted based on Stable Diffusion v1.5, which uses the CLIP ViT-L/14 model by OpenAI as the text encoder. Training images are resized and center-cropped to a resolution of 512×512. For comparison, we use the official implementations of Textual Inversion, DreamBooth, and Custom Diffusion from the Hugging Face diffusers library\footnote{\url{https://github.com/huggingface/diffusers}}, and the official GitHub repository\footnote{\url{https://github.com/forchchch/DisenBooth}} for DisenBooth. Training is performed using the AdamW~\cite{adamw} optimizer with a weight decay of 0.01 on a single NVIDIA RTX 4090 GPU. The batch size is set to 4. The learning rates are as follows: 5e-4 for Textual Inversion, 5e-6 for DreamBooth, 1e-4 for Custom Diffusion, and for our method, 5e-4 for contrastive inversion and 5e-6 for cross-attention fine-tuning. The training steps are set to 3000 for Textual Inversion, 150 for DreamBooth, 300 for Custom Diffusion, and 2000 and 150 for our two-stage method, respectively. During inference, we use the DDIM~\cite{ddim} sampler with 50 steps and apply classifier-free guidance~\cite{cfg} with a scale of 5 to better reflect the text prompt.
\section{Additional Results}

\subsection{Visualization of Learned Auxiliary Tokens} 
\begin{figure*}
  \centering
  \includegraphics[width=1.0\linewidth]{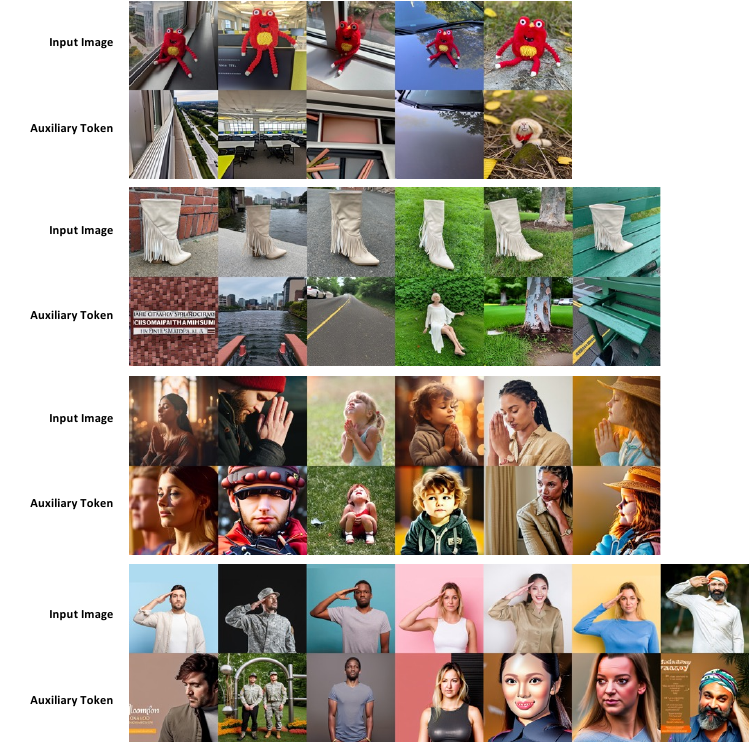}
  \caption{\textbf{Visualization of concepts learned by auxiliary tokens $A_i^*$.} Each column displays an input image (top) and a generated image (bottom) corresponding to is learned auxiliary token. The generated images, prompted by "A photo of $A_i^*$", reveal that the auxiliary tokens capture image-specific context, such as background scenery and composition, while successfully excluding the common target concept. This demonstrates the effectiveness of our contrastive disentanglement.}
  \label{fig:aux_tokens}
\end{figure*}

To verify that our contrastive inversion effectively isolates image-specific features at the token level, we visualize the concepts learned by the auxiliary tokens $A_i^*$. The images in Figure~\ref{fig:aux_tokens} are generated immediately after our first stage (Contrastive Inversion), using the original pre-trained U-Net without any disentangled cross-attention fine-tuning. This demonstrates that before fine-tuning the generator, our method successfully encodes what concept to learn (target) and what to filter out (auxiliary) directly into the token embeddings. The results show that each auxiliary token captures the unique context of its corresponding input image—such as background, object composition, and lighting—while excluding the common target concept. For instance, in the first row, the auxiliary tokens learn concepts like ``a window frame" or ``an office interior" but not the "red monster toy" itself. This confirms that effective concept disentanglement is achieved at the inversion stage.

\subsection{Failure Cases}
\begin{figure*}
  \centering
  \includegraphics[width=1.0\linewidth]{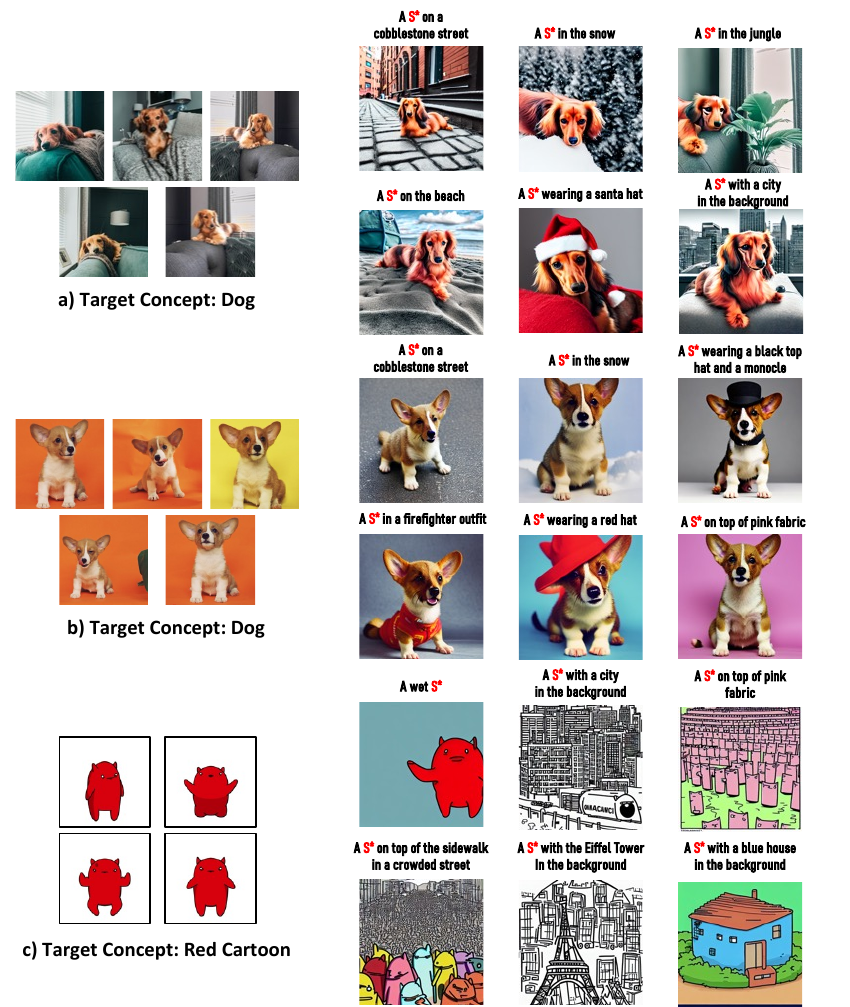}
  \caption{\textbf{Analysis of failure cases.} Our method can fail when the input image set lacks sufficient diversity or distinctive auxiliary information. \textbf{(a, b)} When inputs are too similar (e.g., consistent poses or backgrounds), unintended features become entangled with the target concept, harming editability. \textbf{(c)} When inputs lack distinct auxiliary information (e.g., simple illustrations), the model struggles to separate the target concept (the character) from its intrinsic style.}
  \label{fig:failure_case}
\end{figure*}

Our method's ability to disentangle concepts hinges on the visual cues available in the input image set. Its performance can be compromised if the inputs contain spurious correlations or lack sufficient variation for contrastive learning. Figure~\ref{fig:failure_case} illustrates these failure modes.
\begin{itemize}
\item \textbf{(a), (b) Spurious Correlations in Inputs:} When irrelevant features are consistently present across all images, the model may mistakenly learn them as part of the target concept. In (a), the dog is always in a "lying down" pose, causing the model to entangle this pose with the dog's identity. As a result, it struggles to generate the dog in other poses. Similarly, in (b), the consistent centered composition and monochromatic background are memorized as part of the target concept, limiting editability.
\item \textbf{(c) Insufficient Auxiliary Information:} Conversely, if the images lack distinct auxiliary features (e.g., simple illustrations on a plain background), the contrastive signal for disentanglement becomes weak. In (c), the auxiliary tokens fail to capture meaningful information, preventing the target token from cleanly isolating the "red character" from its ``illustrative style." The model thus learns the style as part of the target concept, rather than just the character's identity.
\end{itemize}

\end{document}